\documentclass[runningheads]{llncs}
\usepackage{graphbox} %
\usepackage{amsmath,amssymb} %
\usepackage{color}
\usepackage[width=122mm,left=12mm,paperwidth=146mm,height=193mm,top=12mm,paperheight=217mm]{geometry}

\newcommand{\isArXiv}[2]{#1}

\pdfoutput=1
\usepackage[numbers,sort,compress]{natbib} %
\usepackage{placeins} %
\usepackage{afterpage}
\usepackage{adjustbox} %
\usepackage{multirow}
\def\eg{\emph{e.g.\ }}
\def\ie{\emph{i.e.\ }}
\def\etc{\emph{etc}}
\newcommand{\etal}{\emph{et al.\@}}
\usepackage[caption=false]{subfig}

\isArXiv{
\usepackage[customdriver=hpdftex,pagebackref=false,breaklinks=true,letterpaper=true,colorlinks=false,bookmarks=false]{hyperref}
\hypersetup{pdfauthor={Relja Arandjelovic},pdftitle={Objects that Sound}}
}{
\usepackage[implicit=false]{hyperref}  %
}

\begin{document}
\pagestyle{headings}
\mainmatter

\title{Objects that Sound}

\titlerunning{Objects that Sound}

\authorrunning{R. Arandjelovi\'c \and A. Zisserman}

\author{Relja Arandjelovi\'c\inst{1} and Andrew Zisserman\inst{1,2}}

\institute{
    DeepMind
	\and
    VGG, Department of Engineering Science, University of Oxford
}

\maketitle

\newcommand{\figTeaser}{
\begin{figure}[t]
\def\teasW{0.25\linewidth}
\centering
    \hfill
    \subfloat[Input image with sound]{
        \centering
        \hspace*{1cm}~
        \includegraphics[width=\teasW]{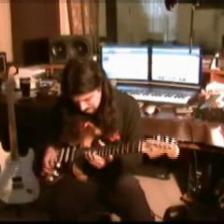}
        ~\hspace*{1cm}
    }
    \subfloat[Where is the sound?]{
        \centering
        \hspace*{1cm}~
        \includegraphics[width=\teasW]{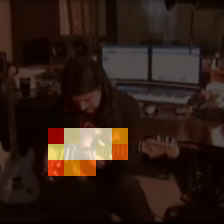}
        ~\hspace*{1cm}
    }
    \hfill
\caption{{\bf Where is the sound?}
Given an input image and sound clip,
our method learns, without a single labelled example, to
localize the object that makes the sound.
}
\label{fig:teaser}
\end{figure}
}

\newcommand{\figRetArch}{
\def\archH{6.5cm}
\begin{figure*}[t]
\captionsetup[subfigure]{font=scriptsize,labelfont=scriptsize}
\centering
    \subfloat[Vision ConvNet]{
        \centering
        \hspace*{0.5cm}~
        \includegraphics[height=\archH]{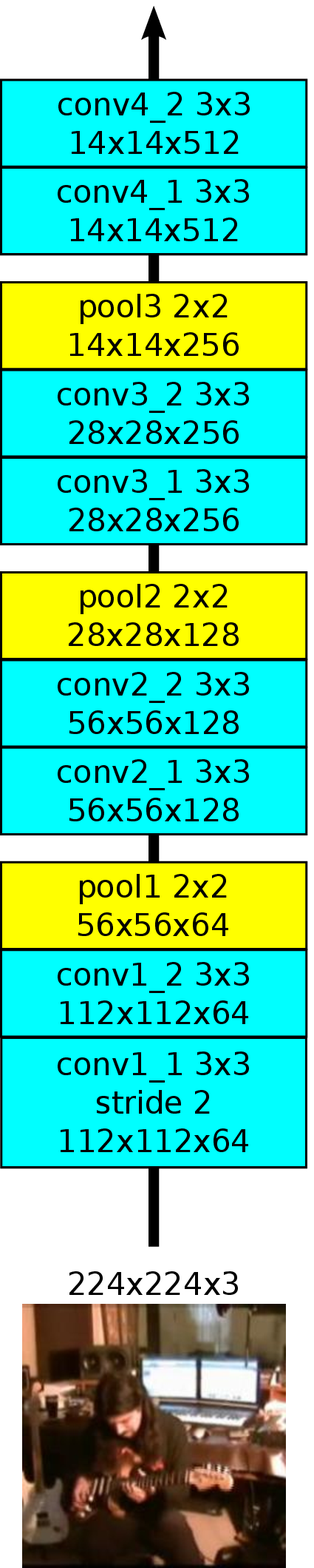}
        ~\hspace*{0.5cm}
        \label{fig:retarch:vision}
    }
    \subfloat[Audio ConvNet]{
        \centering
        \includegraphics[height=\archH]{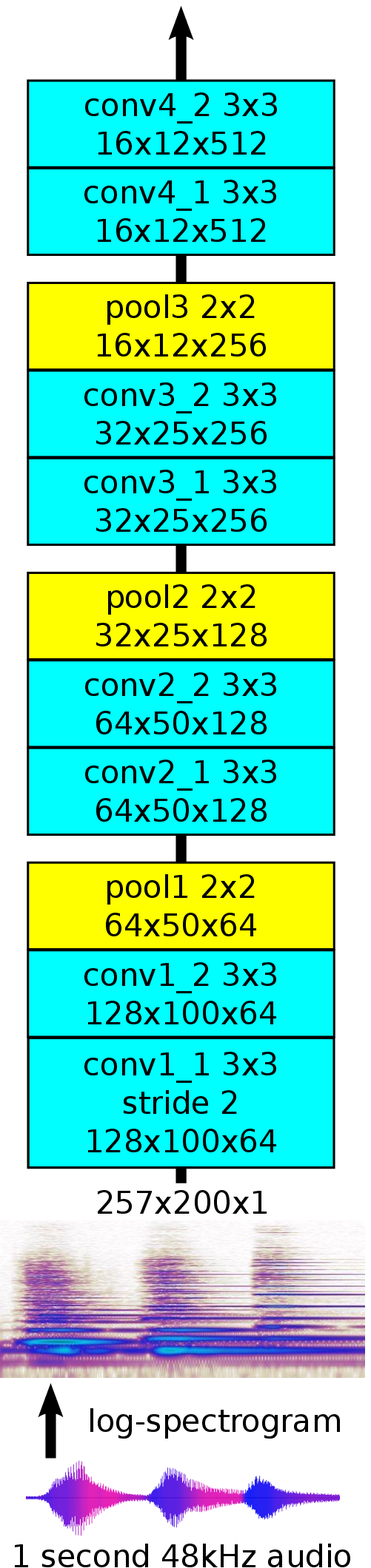}
        ~\hspace*{0.5cm}
        \label{fig:retarch:audio}
    }
    \subfloat[AVE-Net]{
        \centering
        \includegraphics[height=\archH]{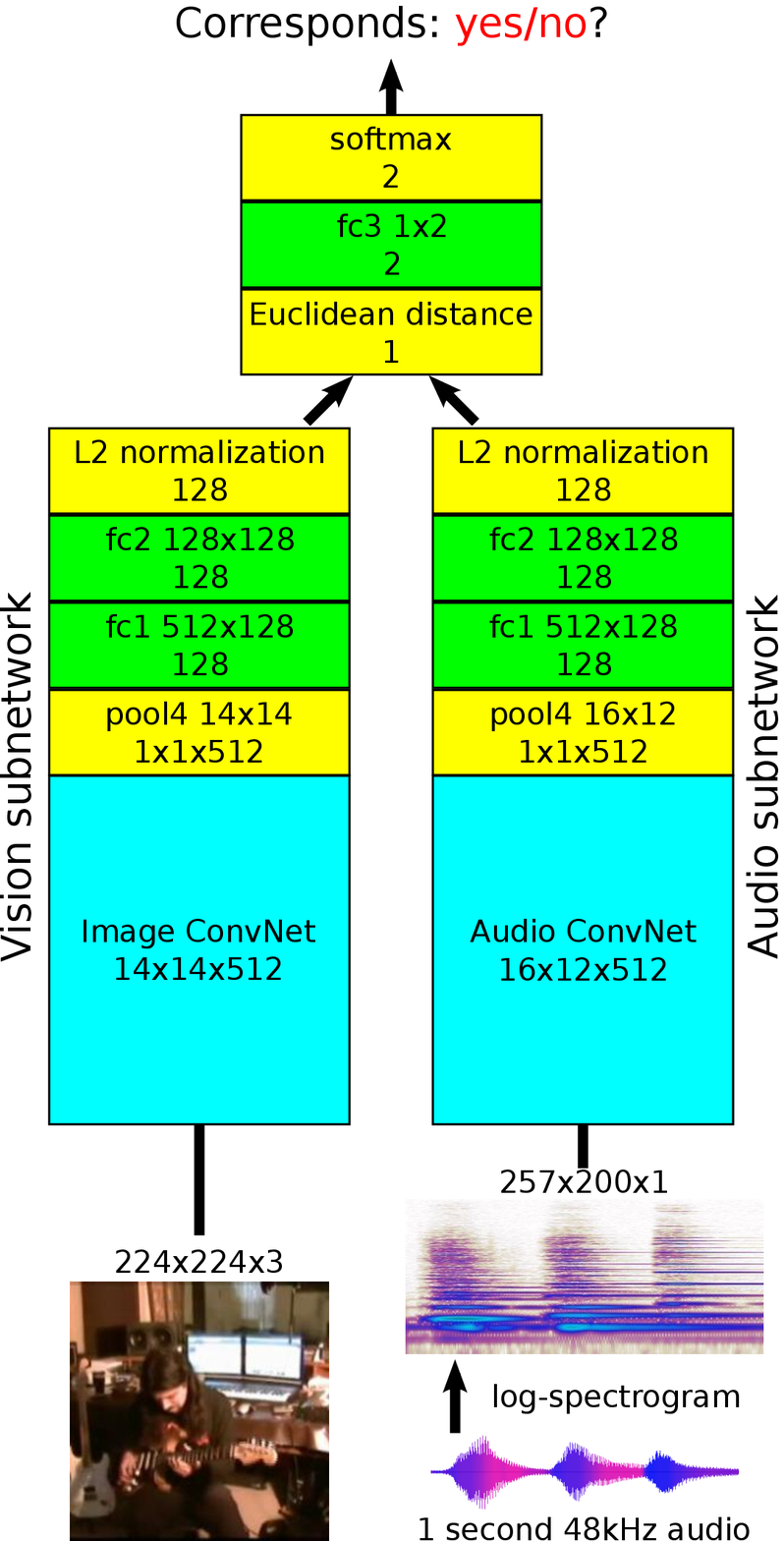}
        \label{fig:retarch:single}
    }
    \subfloat[$L^3$-Net~\cite{Arandjelovic17}]{
        \centering
        \includegraphics[height=\archH]{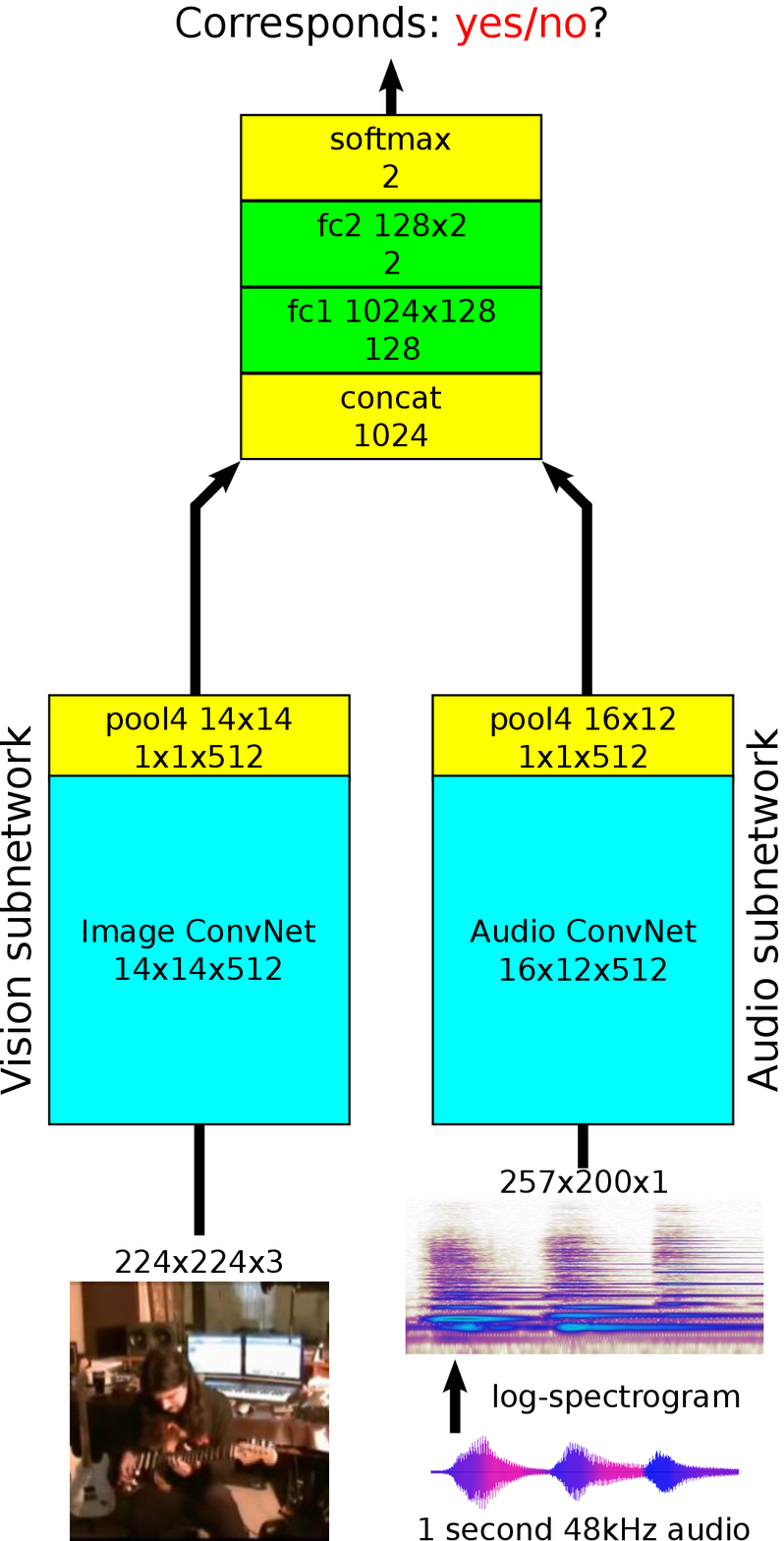}
        \label{fig:retarch:l3}
    }
\caption{{\bf ConvNet architectures}.
Each blocks represents a single layer with text providing more information --
first row: layer name and optional kernel size,
second row: output feature map size.
Each convolutional layer is followed by batch normalization~\cite{Ioffe15}
and a ReLU nonlinearity, and the first fully connected layer (\texttt{fc1})
is followed by ReLU.
All pool layers perform max pooling and their strides are equal to the kernel sizes.
(a) and (b) show the vision and audio ConvNets which perform initial
feature extraction from the image and audio inputs, respectively.
(c) Our AVE-Net is designed to produce aligned vision and audio embeddings
as the only information, a single scalar, used to decide whether the
two inputs correspond is the Euclidean distance between the embeddings.
(d) In contrast, the $L^3$-Net~\cite{Arandjelovic17} architecture combines the two
modalities by concatenation and a couple of fully connected layers
which produce the corresponds or not classification scores.
}
\label{fig:retarch}
\end{figure*}
}

\newcommand{\figMILArch}{
\begin{figure}[t]
\centering
    \includegraphics[height=8cm]{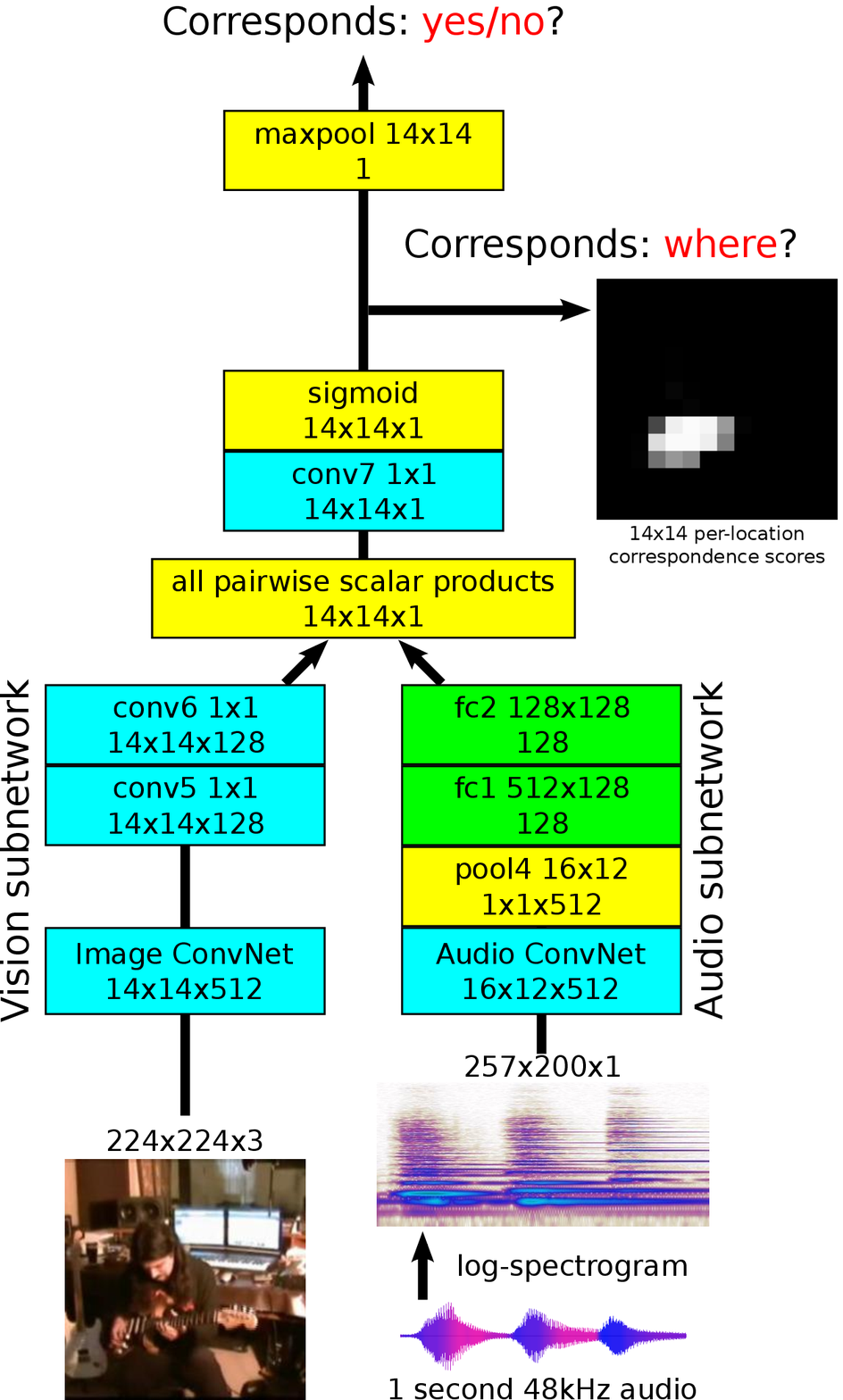}
\caption{{\bf Audio-Visual Object Localization (AVOL-Net).}
The notation and some building blocks are shared with Figure~\ref{fig:retarch}.
The audio subnetwork is the same as in AVE-Net
(Figure~\ref{fig:retarch:single}).
The vision network, instead of globally pooling the feature tensor,
continues to operate at the $14 \times 14$ resolution, with relevant FCs
(vision-\texttt{fc1}, vision-\texttt{fc2}, \texttt{fc3}) converted into
their ``fully convolutional'' equivalents
(\ie $1 \times 1$ convolutions \texttt{conv5}, \texttt{conv6}, \texttt{conv7}).
The similarities between the audio and all vision embeddings
reveal the location of the object that makes the sound, while the maximal
similarity is used as the correspondence score.
}
\label{fig:milarch}
\end{figure}
}

\newcommand{\tabRetrieval}{
\begin{table}[t]
    \centering
\caption{{\bf Cross-modal and intra-modal retrieval.}
Comparison of our method with unsupervised and supervised baselines in terms
of the average nDCG@30 on the AudioSet-Instruments test set.
The columns headers denote the modalities of the query and the database,
respectively,
where \emph{im} stands for \emph{image} and \emph{aud} for \emph{audio}.
Our AVE-Net beats all baselines convincingly.
}
    \small
    \begin{tabular}{l@{~}c@{~~}c@{~~}c@{~~}c}
        Method & im-im & im-aud & aud-im & aud-aud \\
        \hline\hline
        Random chance & .407 & .407 & .407 & .407 \\
        $L^3$-Net~\cite{Arandjelovic17} & .567 & .418 & .385 & .653 \\  %
        $L^3$-Net with CCA & .578 & .531 & .560 & .649 \\  %
        VGG16-ImageNet~\cite{Simonyan15} & .600 & -- & -- & -- \\  %
        VGG16-ImageNet + $L^3$-Audio CCA  & .493 & .458 & .464 & .618 \\  %
        AVE-Net & {\bf .604} & {\bf .561} & {\bf .587} & {\bf .665} \\  %
    \end{tabular}
\label{tab:retrieval}
\end{table}
}

\newcommand{\figRetrieval}{
\begin{figure}[t]
\centering
\def\retW{0.11\linewidth}
  \begin{tabular}{lcccccccc}
  \rotatebox[origin=c]{90}{Query} &
  \includegraphics[width=\retW,align=c]{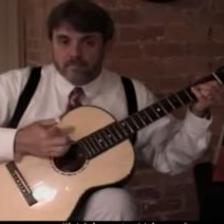} &
  \includegraphics[width=\retW,align=c]{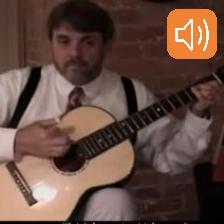} &
  \includegraphics[width=\retW,align=c]{figures/retrieval/sound_401_query.jpg} &
  \includegraphics[width=\retW,align=c]{figures/retrieval/401_query.jpg} &
  \includegraphics[width=\retW,align=c]{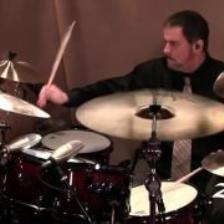} &
  \includegraphics[width=\retW,align=c]{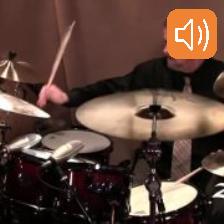} &
  \includegraphics[width=\retW,align=c]{figures/retrieval/sound_2127_query.jpg} &
  \includegraphics[width=\retW,align=c]{figures/retrieval/2127_query.jpg}
  \\[0.7cm]
  \rotatebox[origin=c]{90}{Top 5 retrieved items} &
  \includegraphics[width=\retW,align=c]{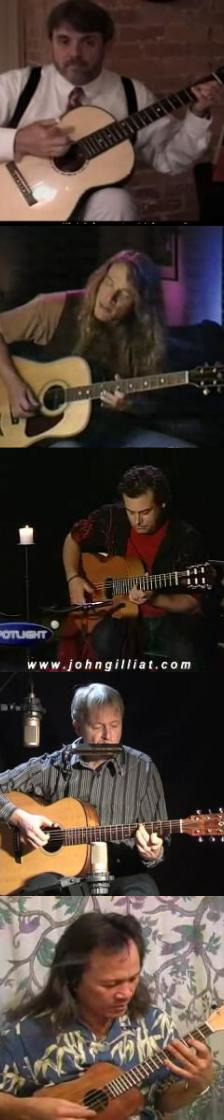} &
  \includegraphics[width=\retW,align=c]{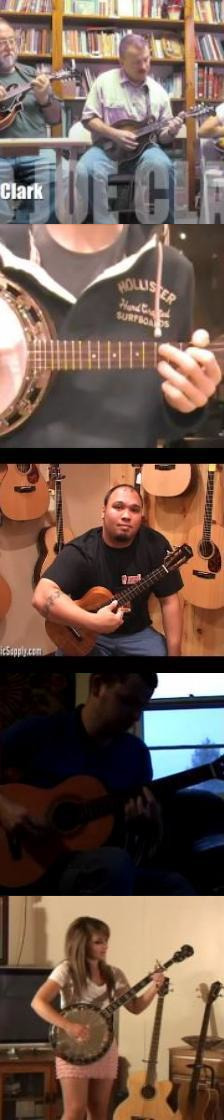} &
  \includegraphics[width=\retW,align=c]{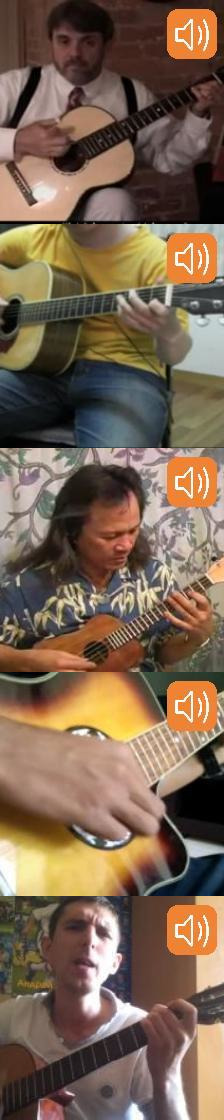} &
  \includegraphics[width=\retW,align=c]{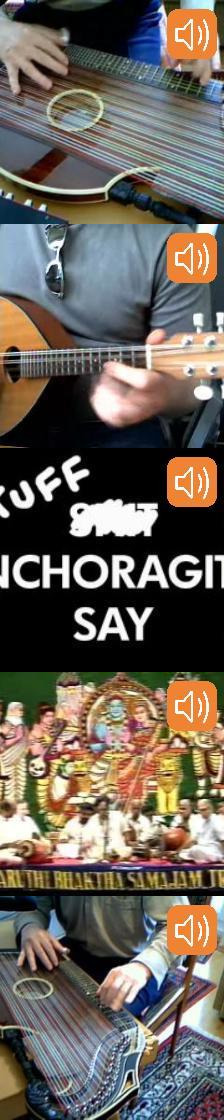} &
  \includegraphics[width=\retW,align=c]{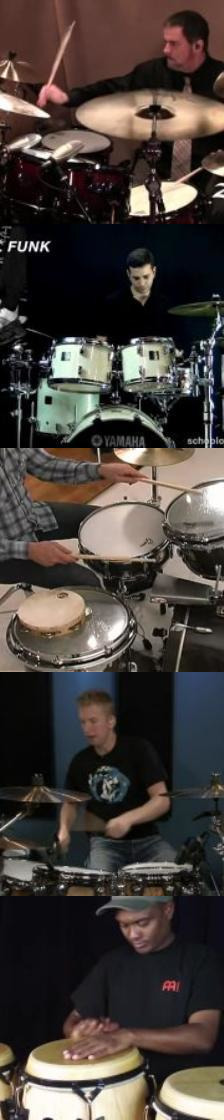} &
  \includegraphics[width=\retW,align=c]{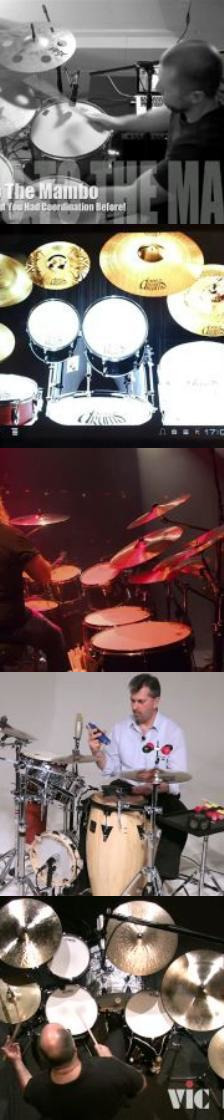} &
  \includegraphics[width=\retW,align=c]{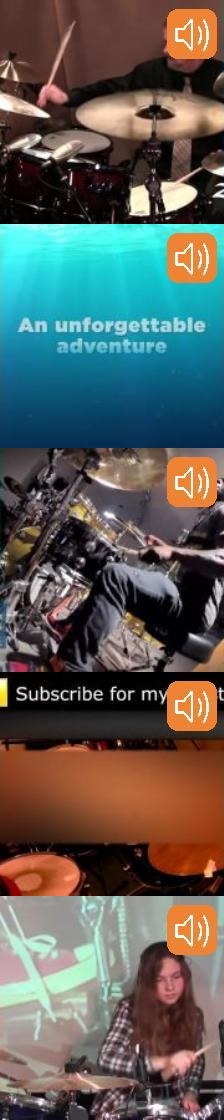} &
  \includegraphics[width=\retW,align=c]{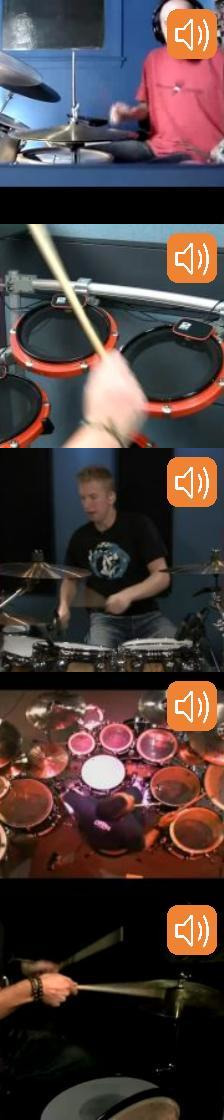}
  \end{tabular}
\caption{{\bf Cross-modal and intra-modal retrieval}.
Each
column
shows one query and retrieved results.
Purely for visualization purposes, as it is hard to display sound,
the frame of the video that is aligned with the sound is shown instead
of the actual sound form.
The sound icon or lack of it indicates the audio or vision modality,
respectively.
For example, the
last column
illustrates query by image into an audio database,
thus answering the question
``Which sounds are the most plausible for this query image?''
Note that many audio retrieval items are indeed correct despite the fact
that their corresponding frames are unrelated --
\eg the audio of the blue image with white text does contain drums --
this is just an artefact of how noisy real-world YouTube videos are.
}
\label{fig:retrieval}
\end{figure}
}

\newcommand{\figLocalization}{
\begin{figure*}[t]
\centering
    \includegraphics[width=0.93\linewidth]{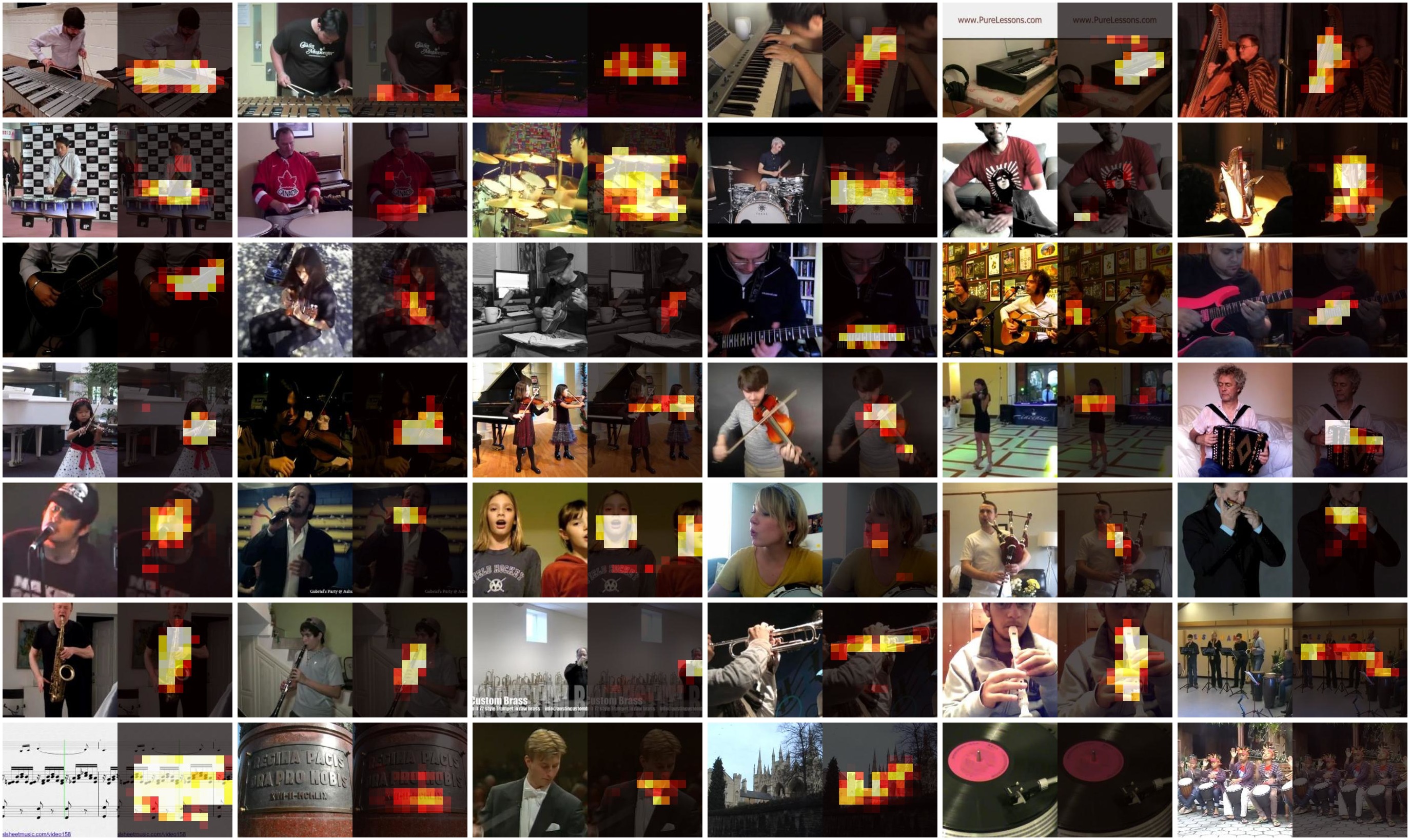}
\caption{{\bf What is making the sound?}
Localization output of the  AVOL-Net on the unseen test data;
see Figure~\ref{fig:teaser} and
\href{https://goo.gl/JVsJ7P}{https://goo.gl/JVsJ7P}
for more.
Recall that the network sees a single frame and therefore cannot ``cheat''
by using motion information.
Each pair of images shows the input frame (left) and the localization output
for the input frame and 1 second of audio around it,
overlaid over the frame (right).
Note the wide range of detectable objects, such as keyboards, accordions, drums,
harps, guitars, violins, xylophones, people's mouths, saxophones, etc.
Sounding objects are detected despite significant clutter and
variations in lighting, scale and viewpoint.
It is also possible to detect multiple relevant objects:
two violins, two people singing, and an orchestra.
The final row shows failure cases, where the first two likely reflects
the noise in the training data as many videos contain just music sheets or
text overlaid with music playing,
in columns 3-4 the network probably just detects the salient parts of the scene,
while in columns 5-6 it fails to detect the sounding objects.
}
\label{fig:localization}
\end{figure*}
}

\newcommand{\figLocalizationTwo}{
\begin{figure*}[t]
\centering
    \includegraphics[width=0.93\linewidth]{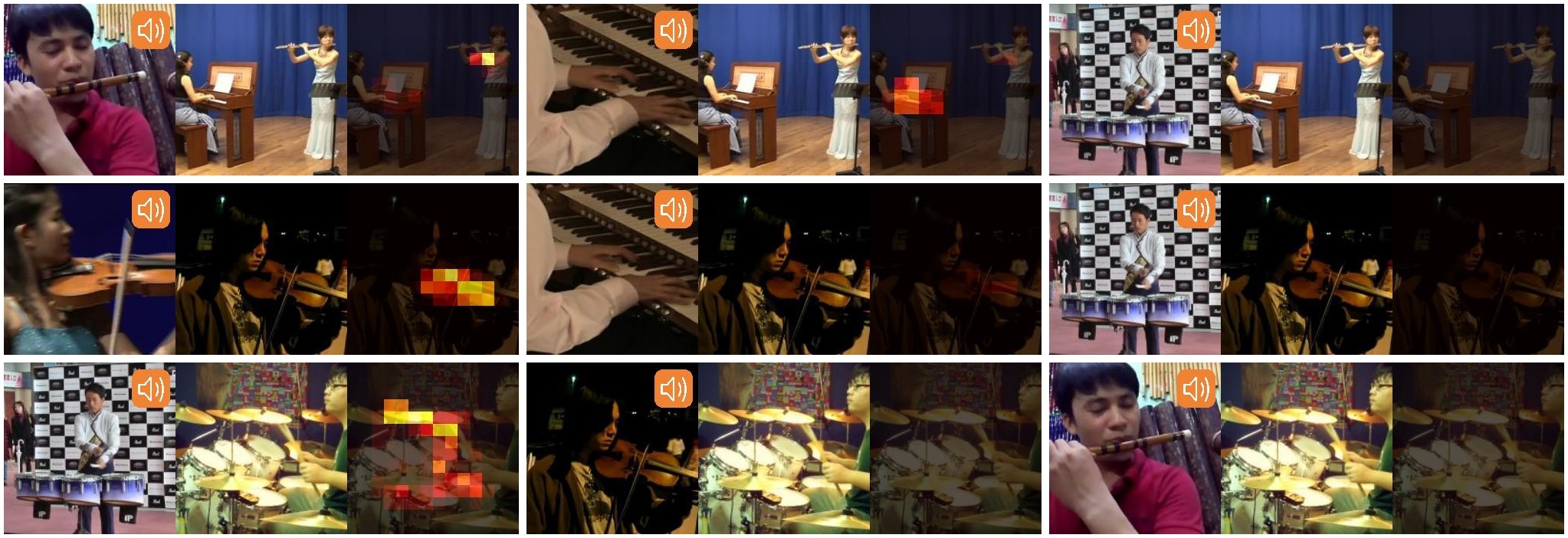}
\caption{{\bf What would make this sound?}
Similarly to Figure~\ref{fig:localization},
the AVOL-Net localization output is shown given an input image frame and 1s of audio.
However, here the frame and audio are mismatched.
Each triplet of images shows the (left) input audio, (middle) input frame,
and (right) localization output overlaid over the frame.
Purely for visualization purposes, as it is hard to display sound,
the frame of the video that is aligned with the sound is shown instead
of the actual sound form (left).
On the example of the first triplet:
(left) flute sound illustrated by an image of a flute,
(middle) image of a piano and a flute,
(right) the flute from the middle image is highlighted as our network
successfully answers the question
``What in the piano-flute image would make a flute sound?''
In each row the input frame is fixed while the input audio varies,
showing that object localization does depend on the sound and therefore
our system is not just detecting salient objects in the scene but
is achieving the original goal -- localizing the object that sounds.
}
\label{fig:localization2}
\end{figure*}
}

\newcommand{\figLocVideo}{
\begin{figure}[t]
\def\vidH{8cm}
\centering
\hfill
        \includegraphics[height=\vidH]{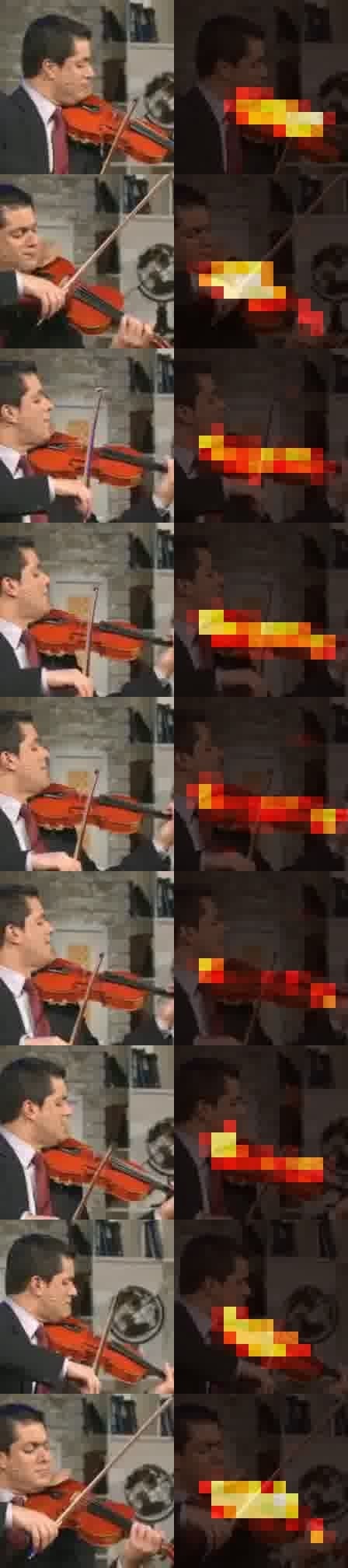}
        \hfill
        \includegraphics[height=\vidH]{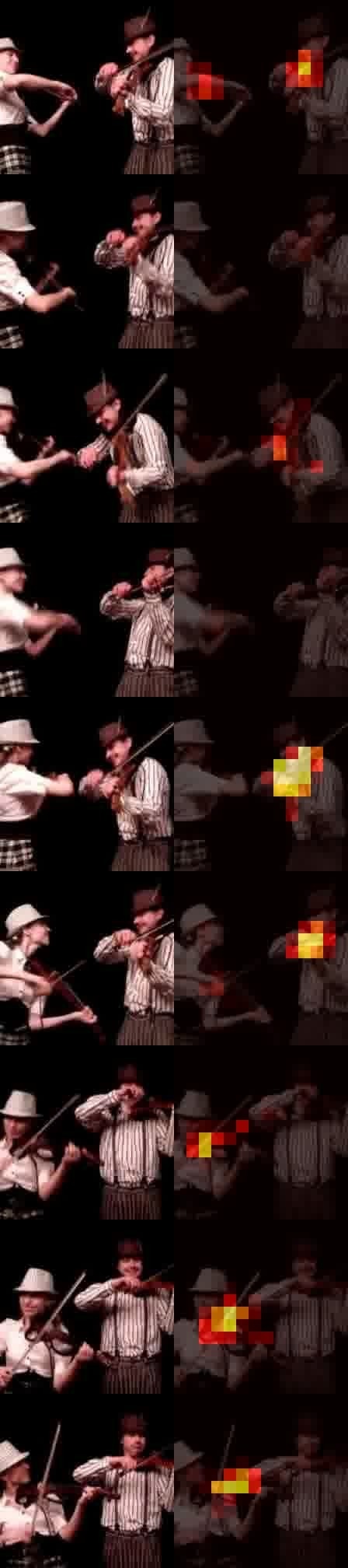}
        \hfill
        \includegraphics[height=\vidH]{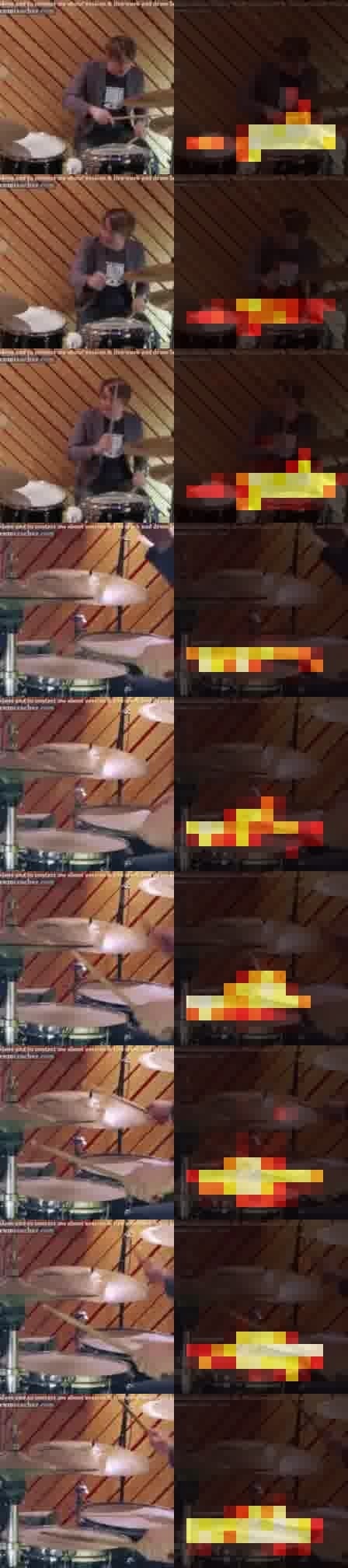}
        \hfill
        \includegraphics[height=\vidH]{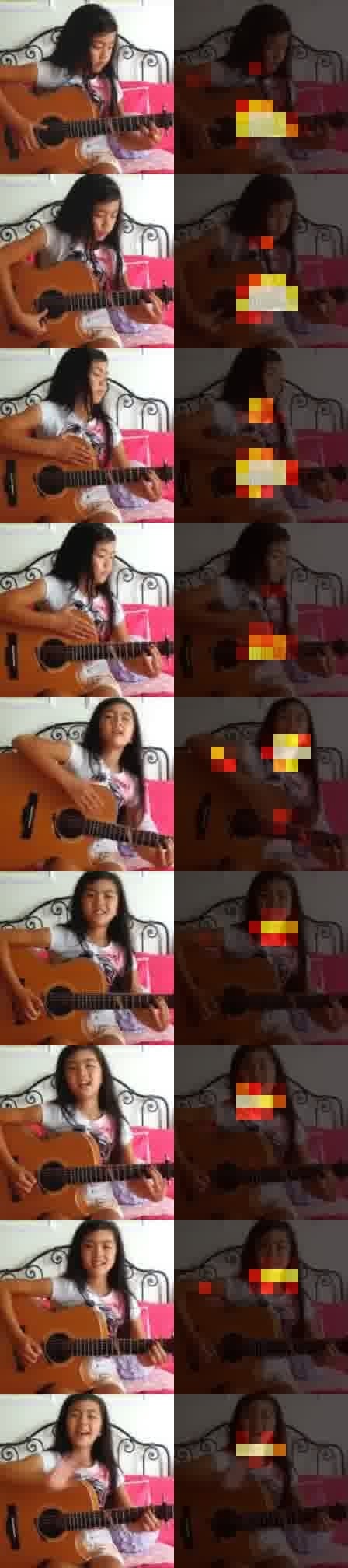}
\hfill~
\caption{{\bf What is making the sound?}
The visualization is the same as for Figure~\ref{fig:localization} but here
each column contains frames from a single video, taken 1 second apart.
The frames are processed completely independently, motion information is
not used, nor there is any temporal smoothing.
Our method reliably detects the sounding object across
varying poses (columns 1-2), and shots (column 3).
Furthermore, it is able to switch between objects that are making the sound
such as interleaved speech and guitar during a guitar lesson (column 4).
}
\label{fig:locvideo}
\end{figure}
}

\begin{abstract}
In this paper our objectives are, first, networks that can embed audio
and visual inputs into a common space that is suitable for
cross-modal retrieval; and second, a network that can localize the
object that sounds in an image, given the audio signal. We achieve
both these objectives by training from unlabelled video using only
\emph{audio-visual correspondence} (AVC) as the objective
function. This is a form of cross-modal self-supervision from video.

To this end, we design new network architectures that can be trained
for cross-modal retrieval and localizing the sound source in an image,
by using the AVC task.
We make the following contributions:  (i) 
show that audio and visual embeddings can be learnt that enable
both within-mode (\eg audio-to-audio) and between-mode retrieval; (ii) explore various architectures
for the AVC task, including those for the visual stream that ingest a single image, or multiple images, or
a single image and multi-frame optical flow;  (iii) show that the semantic object that sounds within an image can
be localized (using only the sound, no motion or flow information); and (iv) give a cautionary tale on 
how to avoid undesirable shortcuts in the data preparation.

\end{abstract}

\section{Introduction}

There has been a recent surge of interest in cross-modal learning from
images and audio~\cite{Aytar16,Harwath16,Owens16a,Arandjelovic17}.  
One reason for this surge is the availability of
virtually unlimited training material in the form of videos (\eg
from YouTube) that can provide both an image stream and  a (synchronized) audio stream,
and this cross-modal information can be used to train deep
networks. Cross-modal learning itself has a long history in computer vision,
principally in the form of images and text~\cite{Barnard02,Duygulu02,Frome13}.  
Although audio and
text share the fact that they are both sequential in nature, the challenges
of using audio to partner images are significantly different to those
of using text.  Text is much closer to a semantic annotation than
audio. With text, \eg in the form of a provided caption of an image,
the concepts (such as `a dog') are directly available and the problem
is then to provide a correspondence between the noun `dog' and a
spatial region in the image~\cite{Barnard02,Xu15}. Whereas, for audio, obtaining the
semantics is less direct, and has more in common with image classification, in that the
concept dog is not directly available from the signal but requires
something like a ConvNet to obtain it (think of classifying an image
as to whether it contains a dog or not, and classifying an audio clip
as to whether it contains the sound of a dog or not).  

\figTeaser

In this paper our interest is in cross-modal learning from images and 
audio~\cite{DeSa94,Kidron05,Owens16,Owens16a,Aytar16,Harwath16,Arandjelovic17,Aytar17}.
In particular, we use unlabelled video as our source material, and employ 
\emph{audio-visual correspondence} (AVC) as the training objective~\cite{Arandjelovic17}.
In brief,  given an input  pair of
a video frame and 1 second of audio, the AVC task requires the  network to decide whether
they are in correspondence or not.
The labels for the positives (matching) and negatives (mismatched) pairs
are obtained directly,  as videos provide an automatic
alignment between the visual and the audio streams --
frame and audio coming from the same time in a video are positives,
while frame and audio coming from different videos are negatives.
As the labels are constructed directly from the data itself,
this is an example of 
``self-supervision''~\cite{Dosovitskiy14,Doersch15,Agrawal15,Wang15,Zhang16,Misra16,Pathak16,Noroozi16,Fernando17,Doersch17}, 
a subclass of unsupervised methods.

The AVC task stimulates the learnt visual and audio representations to be
both discriminative, to distinguish between matched and mismatched pairs,
and semantically meaningful.
The latter is the case because
the only way for a network to solve the task is if it learns to classify
semantic concepts in both modalities, and then judge whether the two concepts
correspond. Recall that the visual network only sees a single frame of video
and therefore it cannot learn to cheat by exploiting motion information.

In this paper we propose two networks that enable new functionalities:
in Section~\ref{sec:retrieval} we propose a network architecture that
produces {\em embeddings} directly suitable for cross-modal retrieval;
in Section~\ref{sec:objectssound}
we design a network and a learning procedure capable of {\em localizing}
the sound source, \ie answering the 
basic question --
``Which object in an image is making the sound?''.  An example is shown in Figure~\ref{fig:teaser}.
Both of these are trained from scratch with no labels whatsoever, using the same
unsupervised audio-visual correspondence task (AVC).

\section{Dataset}
\label{sec:dataset}

Throughout the paper we use the publicly available AudioSet
dataset~\cite{Gemmeke17}. It consists of 10 second clips from YouTube with
an emphasis on audio events, and video-level audio class labels
(potentially more than 1 per video) are available, but are noisy;
the labels are organized in an ontology.
To make the dataset more manageable and interesting for our purposes,
we filter it for sounds of musical instruments, singing and tools,
yielding 110 audio classes
(the full list is given in
\isArXiv{Appendix~\ref{sec:datasetclasses}}{the appendix \cite{Arandjelovic17b}}),
removing uninteresting classes like
breathing, sine wave, sound effect, infrasound, silence, \etc.
The videos are challenging as many are of poor quality,
the audio source is not always
visible, and the audio stream can be artificially inserted on top of the video,
\eg it is often the case that a video is compiled of a musical piece
and an album cover, text naming the song, still frame of the musician,
or even completely unrelated visual motifs like a landscape, \etc.
The dataset already comes with a public train-test split, and we randomly
split the public training set into training and validation sets in
90\%-10\% proportions.
The final {\em AudioSet-Instruments} dataset contains 263k, 30k and 4.3k 10 s clips
in the train, val and test splits, respectively.

We re-emphasise that no labels whatsoever are used for any of our methods
since we treat the dataset purely as a collection of label-less videos.
Labels are only used for quantitative evaluation purposes, \eg to evaluate
the quality of our unsupervised cross-modal retrieval (Section~\ref{sec:retresults}).

\section{Cross-modal retrieval}
\label{sec:retrieval}

In this section we describe a network architecture capable of learning good
visual and audio embeddings from scratch and without labels.
Furthermore, the two embeddings are  aligned in order to enable
querying across modalities, \eg using an image to search for related sounds.

\figRetArch

The  \emph{Audio-Visual Embedding Network (AVE-Net)}
is designed explicitly to facilitate cross-modal
retrieval. 
The input image and 1 second of audio (represented as a log-spectrogram)
are processed by vision and audio subnetworks
(Figures~\ref{fig:retarch:vision} and~\ref{fig:retarch:audio}),
respectively, followed by feature fusion whose goal is to determine whether
the image and the audio correspond under the AVC task.
The architecture is shown in full detail in
Figure~\ref{fig:retarch:single}.
To enforce feature alignment, the AVE-Net computes
the correspondence score as a function of
the Euclidean distance between the normalized visual and audio embeddings.
This information bottleneck, the single scalar value that summarizes
whether the image and the audio correspond, forces the two embeddings
to be aligned. Furthermore, the use of the Euclidean distance during
training is crucial as it makes the features ``aware'' of the distance metric,
therefore making them amenable to retrieval~\cite{Arandjelovic17a}.

The two subnetworks 
produce  a 128-D L2 normalized  embedding for each of the modalities.
The Euclidean distance between the two 128-D features is computed,
and this single scalar is passed through a tiny FC,
which scales and shifts the distance to calibrate it for the subsequent  softmax.
The bias of the FC essentially learns the threshold on the distance
above which the two features are deemed not to correspond.

\paragraph{Relation to previous works.}
The $L^3$-Net introduced in~\cite{Arandjelovic17} and shown in Figure~\ref{fig:retarch:l3}, was
also trained using the AVC task. However, the $L^3$-Net 
audio and visual features are inadequate for cross-modal retrieval
(as will be shown in the results of Section~\ref{sec:retresults})
as they are not aligned in any way --
the fusion is performed by concatenating the features and
the correspondence score is computed
only {\em after} the fully connected layers.
In contrast,  the AVE-Net moves the fully connected layers into the 
vision and audio subnetworks and directly optimizes the features for cross-modal retrieval.

The training bears resemblance to metric learning via
the contrastive loss~\cite{Chopra05}, but
(i) unlike contrastive loss which requires tuning of the margin
hyper-parameter, ours is parameter-free, and
(ii) it explicitly computes the corresponds-or-not output,  thus making it
directly comparable to the $L^3$-Net while contrastive loss
would require another hyper-parameter for the distance threshold.
Wang \etal~\cite{Wang16} also train a network for cross-modal retrieval
but use a triplet loss which also contains the margin hyper-parameter,
they use pretrained networks, and consider different modalities (image-text)
with fully supervised correspondence labels.
In concurrent work, Hong \etal~\cite{Hong18} use a similar technique
with pretrained networks and triplet loss for joint embedding of music and video.
Recent work of~\cite{Aytar17} also trains networks for cross-modal retrieval,
but uses an ImageNet pretrained network as a teacher. In our case, we train 
the entire network from scratch.
\subsection{Evaluation and results}
\label{sec:retresults}

The architectures are trained on the AudioSet-Instruments train-val set, and evaluated
on the AudioSet-Instruments test set described in Section~\ref{sec:dataset}. Implementation details
are given below in Section~\ref{sec:implementation}.

On the audio-visual correspondence task, AVE-Net achieves
an accuracy of 81.9\%, beating slightly the $L^3$-Net which gets 80.8\%.
However, AVC performance is not the ultimate goal since the task is only
used as a proxy for learning good embeddings,
so the real test of interest here is
the retrieval performance.

To evaluate the intra-modal (\eg image-to-image) and cross-modal retrieval,
we use the AudioSet-Instruments test dataset.
A single frame and surrounding 1 second of audio are sampled randomly
from each test video to form the retrieval database.
All combinations of image/audio as query and image/audio as database
are tested,
\eg audio-to-image uses the {\em audio} embedding as the query vector
to search the database of visual embeddings, answering the question
``Which image could make this sound?''; and
image-to-image uses the {\em visual} embedding as the query vector
to search the same database.

\paragraph{Evaluation metric.}
The performance of a retrieval system is assessed using a standard measure
-- the normalized discounted cumulative gain (nDCG).
It measures the quality of the ranked list of the top $k$ retrieved items
(we use $k=30$ throughout) normalized to the $[0, 1]$ range,
where $1$ signifies a perfect ranking in which items are sorted in
a non-increasing relevance-to-query order.
For details on the definition of the relevance, refer to
\isArXiv{Appendix~\ref{sec:relevance}}{the appendix \cite{Arandjelovic17b}}.
Each item in the test dataset is used as a query and the average
nDCG@30 is reported as the final retrieval performance.
Recall that the labels are noisy,  and note that we only extract a single
frame / 1s audio per video and can therefore miss the relevant event,
so the ideal nDCG of $1$ is highly unlikely to be achievable.

\paragraph{Baselines.}
We compare to the $L^3$-Net as it is also trained in an unsupervised manner,
and we train it using an identical procedure and training data to our method.
As the $L^3$-Net is expected not to work for cross-modal retrieval
since the representation are not aligned in any way, we also test the
$L^3$-Net representations aligned with CCA as a baseline.
In addition, vision features extracted from the last hidden layer
of the VGG-16 network
trained in a fully-supervised manner on ImageNet~\cite{Simonyan15}
are evaluated as well.
For cross-modal retrieval, the VGG16-ImageNet visual features
are aligned with the $L^3$-Net audio features using CCA, which is a strong
baseline as the vision features are fully-supervised while the audio
features are state-of-the-art~\cite{Arandjelovic17}.
Note that the vanilla $L^3$-Net produces 512-D representations,
while VGG16 yields a 4096-D visual descriptor.
For computational reasons, and for fair comparison with our AVE-Net which
produces 128-D embeddings, all CCA-based methods use 128 components.
For all cases the representations are L2-normalized as we found this to
significantly improve the performance; note that AVE-Net includes
L2-normalization in the architecture and therefore the re-normalization
is redundant.

\paragraph{Results.}
The nDCG@30 for all combinations of query-database modalities is shown in
Table~\ref{tab:retrieval}.
For intra-modal retrieval (image-image, audio-audio) our AVE-Net is better than
all baselines including slightly beating VGG16-ImageNet for image-image,
which was trained in a fully supervised manner on another task.
It is interesting to note that our network has never seen same-modality pairs
during training, so it has not been trained explicitly for
image-image and audio-audio retrieval. However, intra-modal retrieval works
because of transitivity -- an image of a violin is close in feature
space to the sound of a violin, which is in turn close to other images of violins.
Note that despite learning essentially the same information on the same
task and training data as the $L^3$-Net, our AVE-Net outperforms the $L^3$-Net
because it is Euclidean distance ``aware'', \ie it has been designed and trained
with retrieval in mind.

For cross-modal retrieval (image-audio, audio-image), AVE-Net
beats all baselines, verifying that our unsupervised training is effective.
The $L^3$-Net representations are clearly not aligned across modalities as
their cross-modal retrieval performance is on the level of random chance.
The $L^3$-Net features aligned with CCA form a strong baseline, but
the benefits of directly training our network for alignment are apparent.
It is interesting that aligning vision features trained on ImageNet
with state-of-the-art $L^3$-Net audio features using CCA performs worse than
other methods, demonstrating a case for unsupervised learning from a more varied dataset, as it is not
sufficient to just use ImageNet-pretrained networks as black-box
feature extractors.

Figure~\ref{fig:retrieval} shows some qualitative retrieval results,
illustrating the efficacy of our approach.
The system generally does retrieve relevant items from the database,
while making reasonable mistakes such as confusing the sound of
a zither with an acoustic guitar.

\tabRetrieval
\figRetrieval
\afterpage{\FloatBarrier}

\subsection{Extending the AVE-Net to multiple frames}
\label{sec:multiframe}
It is also interesting to investigate whether using information from
multiple frames can help solving the AVC task.
For these results only, we evaluate two modifications to the architecture from
Figure~\ref{fig:retarch:vision} to handle a different visual input --
multiple frames (AVE+MF) and optical flow (AVE+OF).
For conciseness, the details of the architectures are explained
in \isArXiv{Appendix~\ref{sec:aveof}}{the appendix \cite{Arandjelovic17b}},
but the overall idea is that
for AVE+MF we input 25 frames and convert convolution layers from 2D to 3D,
while for AVE+OF we combine information from a single frame and 10 frames of
optical flow using a two-stream network in the style of~\cite{Simonyan14b}.

The performance of the AVE+MF and AVE+OF networks on the AVC task are
84.7\% and 84.9\%, respectively, compared to our single input image
network's 81.9\%.
However, when evaluated on retrieval, they fail to provide a boost,
\eg the AVE+OF network achieves
0.608, 0.558, 0.588, and 0.665 for im-im, im-aud, aud-im and aud-aud,
respectively;
this is comparable to the performance of
the vanilla AVE-Net that uses a single frame as input
(Table~\ref{tab:retrieval}).
One explanation of this underwhelming result is that,
as is the case with most unsupervised approaches,
the performance on the training objective is not necessarily in
perfect correlation with the quality of learnt features and their
performance on the task of interest.
More specifically, the AVE+MF and AVE+OF could be using the motion
information available at input to solve the AVC task more easily by
exploiting some lower-level information
(\eg changes in the  motion could be correlated with changes in sound, such as when seeing the fingers playing a guitar or flute),
which in turn
provides less incentive for the network to learn good semantic
embeddings.
For this reason, a single frame input is used for all other experiments.

\subsection{Preventing shortcuts and Implementation}
\label{sec:implementation}
\paragraph{Preventing shortcuts.}
Deep neural networks are notorious for finding subtle data shortcuts to exploit  in order
to ``cheat'' and thus not learn to solve the task in the desired manner;
an example is the misuse of chromatic aberration in~\cite{Doersch15} to solve the relative-position task.
To prevent such behaviour,
we found it important to carefully implement the sampling of
AVC negative pairs to be as similar as possible to the sampling of positive pairs.
In detail, a positive pair is generated 
by sampling a random video,
picking a random frame in that video, and then picking a
1 second audio with the frame at its mid-point. It is tempting to generate
a negative pair by randomly sampling two
different videos and picking a random frame from one and a random 1 second
audio clip from the other. However, this produces a slight statistical
difference between positive and negative audio samples, in that
the mid-point of the positives is always aligned with a frame and is thus
at a multiple of 0.04 seconds (the video frame rate is 25fps),
while negatives have no such restrictions. 
This allows a shortcut as
it appears the network is able to learn to recognize audio samples
taken at multiples of 0.04s, therefore distinguishing
positives from negatives.
It probably does so by exploiting
low-level artefacts of MPEG encoding and/or audio resampling.
Therefore, with this naive implementation of negative pair generation the network has less
incentive to strongly learn semantically meaningful information.

To prevent this from happening, the audio for the negative pair is also
sampled only from multiples of 0.04s.
Without shortcut prevention, the AVE-Net achieves
an artificially high accuracy of 87.6\% on the AVC task,
compared to 81.9\% with the proper sampling safety mechanism in place,
but the performance of the
network without shortcut prevention
on the retrieval task is consistently 1-2\% worse.
Note that, for fairness, we train the $L^3$-Net with shortcut prevention
as well.

The $L^3$-Net training in~\cite{Arandjelovic17} does not encounter this problem
due to performing additional data augmentation by randomly misaligning
the audio and the frame by up to 1 second for both positives and negatives.
We apply this  augmentation as well,
but our observation is important to keep in mind for future unsupervised
approaches where exact alignment might be required, such as
audio-visual synchronization.

\paragraph{Implementation details.}
We follow the same setup and implementation details as in~\cite{Arandjelovic17}.
Namely, the input frame is a $224 \times 224$ colour image,
while the 1 second of audio is resampled at 48 kHz, converted into
a log-spectrogram (window length 0.01s and half-window overlap)
and treated as a $257 \times 200$ greyscale image.
Standard data augmentation is used -- random cropping, horizontal flipping and
brightness and saturation jittering for vision, and random clip-level amplitude
jittering for audio.
The network is trained with cross-entropy loss for the binary classification
task -- whether the image and the audio correspond or not --
using the Adam optimizer~\cite{Kingma15}, weight decay $10^{-5}$,
and learning rate obtained by grid search.
Training is done using 16 GPUs in parallel with synchronous updates
implemented in TensorFlow, where each worker processes a 128-element batch,
thus making the effective batch size 2048.

Note that the only small differences from the setup of~\cite{Arandjelovic17}
are that:
(i) We use a stride of 2 pixels in the first convolutional layers as we found it
to not affect the performance
while yielding a $4 \times$
speedup and saving in GPU memory, thus enabling the use of $4 \times$
larger batches (the extra factor of $2 \times$ is through use of a better GPU);
and
(ii) We use a learning rate schedule in the style of~\cite{Szegedy15} where
the learning rate is decreased by 6\% every 16 epochs.
With this setup we are able to fully reproduce the $L^3$-Net results of~\cite{Arandjelovic17},
achieving even slightly better performance
(+0.5\% on the ESC-50 classification benchmark~\cite{Piczak15}),
probably due to the improved
learning rate schedule and the use of larger batches.

\section{Localizing objects that sound}
\label{sec:objectssound}

A system which understands the audio-visual world should associate appearance
of an object with the sound it makes, and thus be able to answer
``where is the object that is making the sound?''
Here we outline an architecture and a training procedure
for learning to localize the sounding object, while still operating
in the scenario where there is no supervision, neither on the
object location level nor on their identities.
We again make use of the AVC task, and
show that by designing the network
appropriately, it is possible to learn to
localize sounding objects in this extremely challenging label-less scenario.

In contrast to the standard AVC task where the goal is to learn a single
embedding of the entire image which explains the sound, the goal in
sound localization is to find  regions  of the image which explain
the sound, while other regions should not be correlated with it and belong
to the background. To operationalize this, we formulate the problem in the
Multiple Instance Learning (MIL) framework~\cite{Dietterich97}.
Namely, local region-level image descriptors are extracted on a spatial grid
and a similarity score is computed between the audio embedding and
each of the vision descriptors.
For the goal of  finding regions  which correlate well with the sound,
the maximal similarity score is used as the measure of the image-audio
agreement. The network is then trained in the same manner as for the
AVC task, \ie  predicting whether the image and the audio
correspond.
For corresponding pairs, the method encourages one region to respond highly
and therefore localize the object, while for mismatched pairs
the maximal score should be low thus making the entire score map low,
indicating, as desired, there is no object which makes the input sound.
In essence, the audio representation forms a filter
which ``looks'' for relevant image patches in a similar manner to an 
attention mechanism.

\figMILArch

Our \emph{Audio-Visual Object Localization Network (AVOL-Net)}
is depicted in Figure~\ref{fig:milarch}.
Compared to the AVE-Net (Figure~\ref{fig:retarch:single}),
the vision subnetwork does not pool \texttt{conv4\_2} features but
keeps operating on the $14 \times 14$ resolution. To enable this,
the two fully connected layers \texttt{fc1} and \texttt{fc2} of the
vision subnetwork are converted to $1 \times 1$ convolutions
\texttt{conv5} and \texttt{conv6}.
Feature normalization is removed to enable features to have a low response
on background regions.
Similarities between each of the $14 \times 14$ 128-D visual descriptors
and the single 128-D audio descriptor are computed via a scalar product,
producing a $14 \times 14$ similarity score map.
Similarly to the AVE-Net, the scores are calibrated using
a tiny $1 \times 1$ convolution
(\texttt{fc3} converted to be ``fully convolutional''),
followed by a sigmoid %
which produces the localization output in the form of
the image-audio correspondence score for each spatial location.
Max pooling over all spatial locations is performed to obtain
the final correspondence score, which is then used for training
on the AVC task using the logistic loss. %

\paragraph{Relation to previous works.}
While usually hinting at object localization, previous cross-modal works
fall short from achieving this goal.
Harwath \etal~\cite{Harwath16} demonstrate localizing objects
in the audio domain of a spoken text, but
do not design their network for localization.
In~\cite{Arandjelovic17}, the network, trained from scratch,
internally learns object detectors, but has never been demonstrated to be
able to answer the question
``Where is the object that is making the sound?'', nor, unlike our approach,
was it trained with this ability in mind.
Rather, their heatmaps are produced by examining responses of its various
neurons given \emph{only} the input image.
The output is computed \emph{completely independently of the sound} and
therefore cannot answer ``Where is the object that is making the sound?''.

Our approach has similarities with~\cite{Oquab15} and~\cite{Zhou16}
who used max and average pooling, respectively,
to learn object detectors without bounding box annotations
in the single visual modality setting,
but use ImageNet pretrained networks and image-level labels.
The MIL-based approach also has connections with attention mechanisms
as it can be viewed as ``infinitely hard'' attention~\cite{Bahdanau15,Xu15}.
Note that we do not use information from multiple audio channels which could
aid localization~\cite{Shivappa10} because
(i) this setup generally requires known calibration of the multi-microphone rig
which is unknown for unconstrained YouTube videos,
(ii) the number of channels changes across videos,
(iii) quality of audio on YouTube
varies significantly while localization methods
based on multi-microphone information are prone to
noise and reverberation, and
(iv) we desire that our system learns to detect semantic concepts
rather than localize by ``cheating'' through accessing
multi-microphone information.
Finally, a similar technique to ours appears in the concurrent work of \cite{Senocak18},
while later works of \cite{Zhao18,Owens18} are also relevant.
\subsection{Evaluation and results}

First, the accuracy of the localization network (AVOL-Net)
on the AVC task is the same
as  that of the AVE-Net embedding network in  Section~\ref{sec:retrieval},
which is encouraging as it means that
switching to the MIL setup does not cause a loss in accuracy and the ability
to detect semantic concepts in the two modalities.

\figLocalization
\figLocalizationTwo

The ability of the network to localize the object(s) that sound is demonstrated
in Figure~\ref{fig:localization}. It is able to detect a wide range of objects
in different viewpoints and scales, and under challenging imaging conditions.
A more detailed discussion including the analysis of some failure cases
is available in the figure caption.
As expected from an unsupervised method, it is not necessarily the case that
it detects the entire object but can focus only on specific discriminative
parts such as the interface between the hands and the piano keyboard.
This interacts with the more philosophical question of what is an object
and what is it that is making the sound --
the body of the piano and its strings, the keyboard, the fingers on the keyboard,
the whole human together with the instrument, or the entire orchestra?
How should a gramophone or a radio be handled by the system, as they can
produce arbitrary sounds?

From the impressive results in Figure~\ref{fig:localization},
one question that comes to mind is whether the network is simply detecting
the salient object in the image, which is not the desired behaviour.
To test this hypothesis we can provide mismatched frame and audio pairs
as inputs to interrogate the network to answer ``what would make this sound?'',
and check if salient objects are still highlighted regardless of the
irrelevant sound. Figure~\ref{fig:localization2} shows that this is indeed
not the case, as when, for example, drums are played on top of an image of
a violin, the localization map is empty. In contrast, when another violin is
played, the network highlights the violin. Furthermore, to completely reject
the saliency hypothesis -- in the case of an image depicting
a piano and a flute, it is possible to play a flute sound and the network
will pick the flute, while if a piano is played, the piano is highlighted
in the image. Therefore, the network has truly learnt to disentangle
multiple
objects in an image and maintain a discriminative embedding for each
of them.

To evaluate the localization performance quantitatively, 500 clips are sampled randomly from 
the validation data and the middle frame annotated with the localization of the instrument producing the sound.
We then compare two methods of predicting the localization (as in~\cite{Oquab15}): first, a baseline method that
always predicts the center of the image; second, the mode of the AVOL-Net heatmap produced by
inputting the sound of the clip. The baseline achieves 57.2\%, whilst AVOL-Net achieves 81.7\%. 
This demonstrates that the AVOL-NET is not simply highlighting the salient object at the center of the image.
Failure cases are mainly due to the problems with the AudioSet dataset described in
Section~\ref{sec:dataset}.
Note, it
is necessary to annotate the data, rather than using a standard benchmark, since datasets such
as PASCAL VOC, COCO, DAVIS, KITTI, do not contain
musical instruments. This also means that off-the-shelf object
detectors for instruments are not available, so could not be used 
to annotate AudioSet frames with bounding boxes.

Finally, Figure~\ref{fig:locvideo}  shows the localization results
on videos. Note that each video frame and surrounding audio are processed
completely independently, so no motion information is used, nor is there 
any temporal smoothing. The results reiterate the ability of the system
to detect an object under a variety of poses, and to highlight
different objects depending on the varying audio context.
Please see
\href{https://goo.gl/JVsJ7P}{this YouTube playlist (https://goo.gl/JVsJ7P)}
for more video results.

\figLocVideo

\section{Conclusions and future work}

We have demonstrated that the unsupervised audio-visual correspondence task
enables, with appropriate network design,
two entirely new functionalities to be learnt: cross-modal
retrieval, and semantic based localization of objects that sound.
The AVE-Net was shown to perform cross-modal retrieval even better than
supervised baselines,
while the AVOL-Net exhibits impressive object localization capabilities.
Potential improvements could include modifying the AVOL-Net to have
an explicit soft attention mechanism, rather than the max-pooling used
currently.

\paragraph{Acknowledgements.} We thank Carl Doersch for useful insights
regarding preventing shortcuts.

\clearpage
\bibliographystyle{splncs}
\bibliography{bib/shortstrings,bib/vgg_local,bib/vgg_other,bib/to_add}

\isArXiv{
\appendix

\newcommand{\figOFArch}{
\def\archOVH{10cm}
\begin{figure}[t]
\centering
    \includegraphics[height=\archOVH]{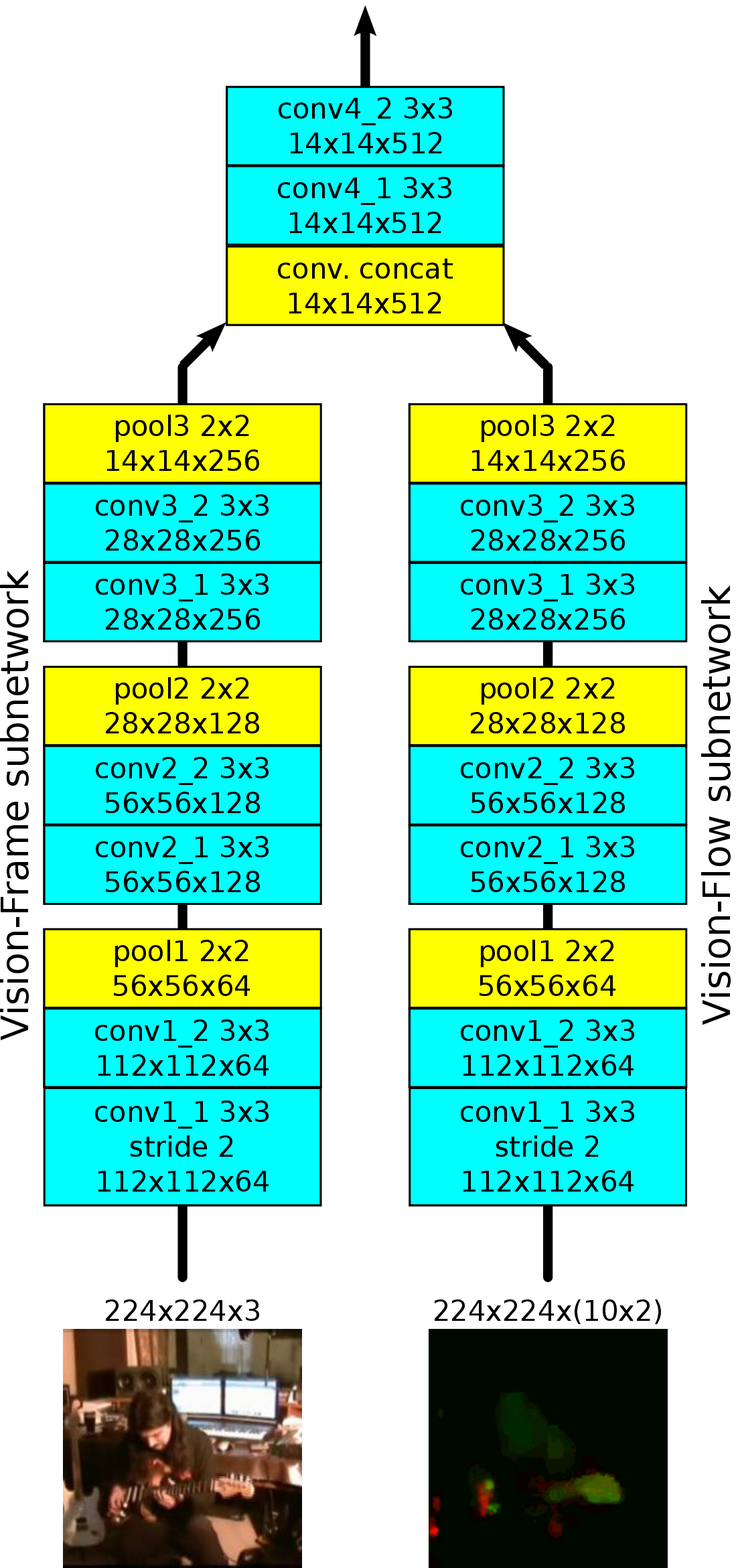}
\caption{{\bf AVE+OF: Vision ConvNet}.
The notation and some building blocks are shared with Figure~\ref{fig:retarch}.
The vision subnetwork of the AVE+OF network is a two-stream
network~\cite{Simonyan14b},
where the image and flow streams are processed independently with
3 \texttt{conv-conv-pool} blocks each, followed by concatenating their outputs
in the `channel` dimension, and passing through another \texttt{conv-conv} block.
The image is a single RGB frame, while there are 10 frames of flow
(concatenated in the `channel` dimension) where each spatial location
contains a 2-D vector of horizontal and vertical displacements.
}
\label{fig:retarch:visionflow}
\end{figure}
}

\newcommand{\figDim}{
\begin{figure}[h]
\centering
        \includegraphics[width=0.55\linewidth]{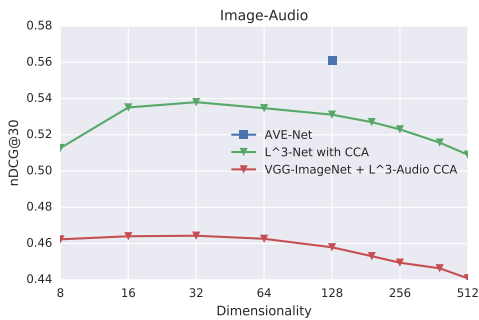}
\caption{{\bf Cross-modal retrieval vs embedding dimensionality}.
Comparison of our method with baselines in terms of the average nDCG@30
on the AudioSet-Instruments test set.
Our AVE-Net beats all baselines regardless for all sizes of baseline embeddings.
Note that CCA does not necessarily work better with increased dimensionality due to denoising properties of dimensionality reduction.
}
\label{fig:ccadim}
\end{figure}
}

\newcommand{\figNdcg}{
\begin{figure}[h]
\def\ndcgW{0.5\linewidth}
\centering
\mbox{
        \includegraphics[width=\ndcgW]{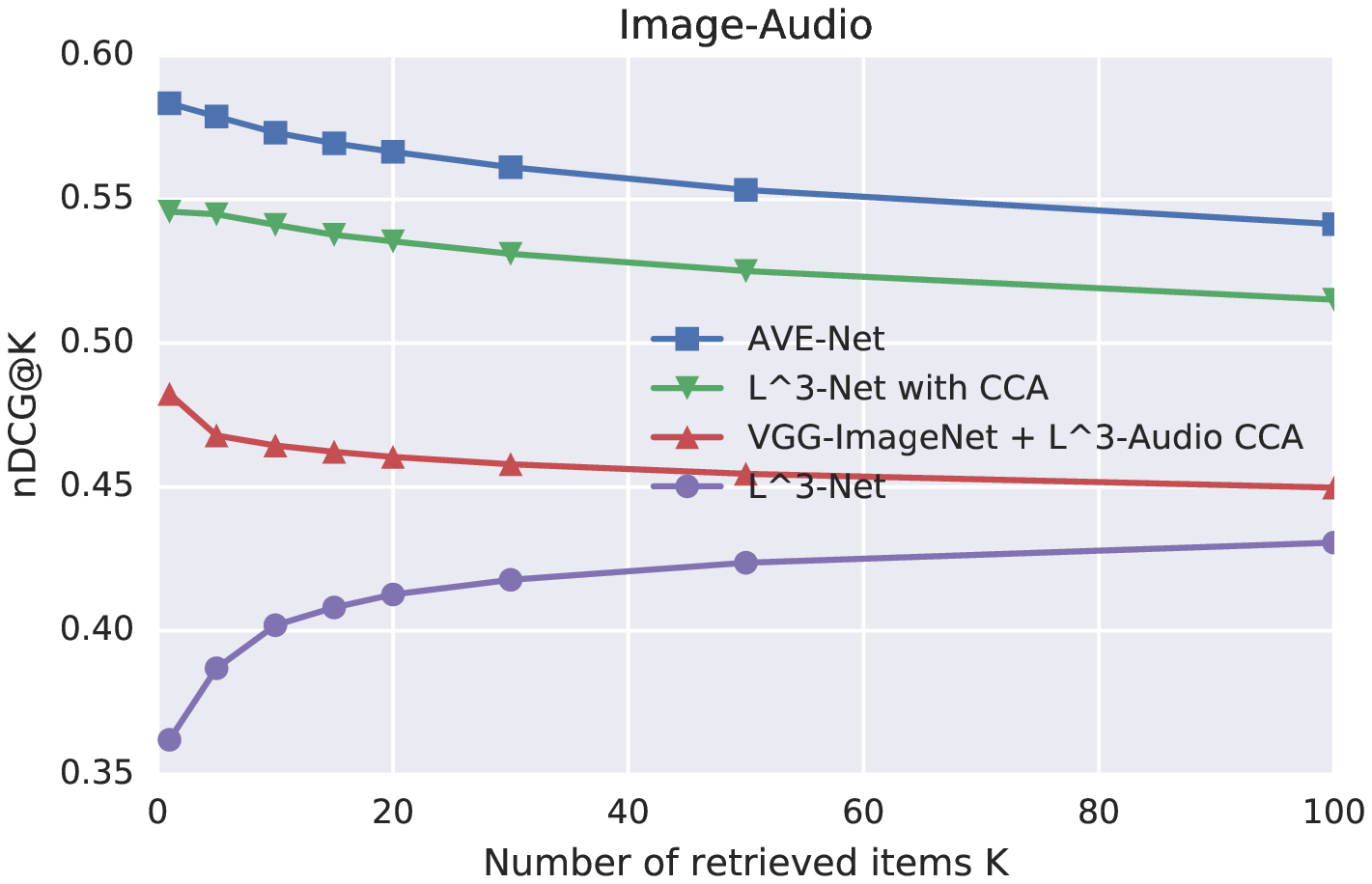}
        \includegraphics[width=\ndcgW]{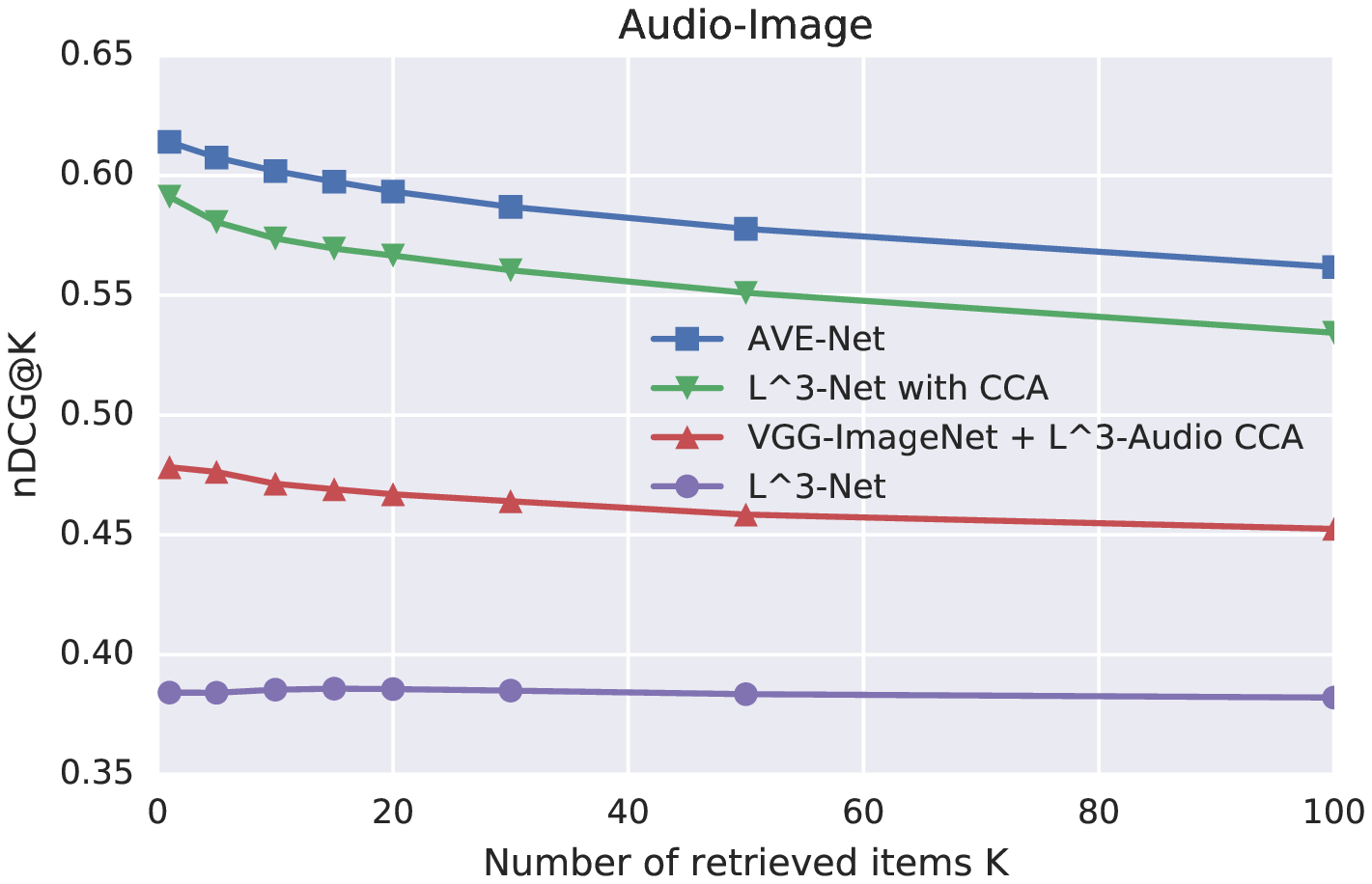}
}
\caption{{\bf Cross-modal retrieval}.
Comparison of our method with baselines in terms of the average nDCG@K
for various values of $K$, on the AudioSet-Instruments test set.
Our AVE-Net beats all baselines for all $K$.
}
\label{fig:ndcgs}
\end{figure}
}

\section{AudioSet-Instruments}

Throughout the paper we use the publicly available AudioSet
dataset~\cite{Gemmeke17} (Section~\ref{sec:dataset}).
It contains video-level audio class labels (potentially more than 1 per video)
which are organized in an ontology;
recall that no labels are used for training, just for evaluation.
This section describes the AudioSet-Instruments subset we use,
and further details needed for evaluation of retrieval performance.

\subsection{Classes}
\label{sec:datasetclasses}

To make the dataset more manageable and interesting for our purposes,
we filter it for sounds of musical instruments, singing and tools,
\ie we use all videos which contain at least one label that is a descendant
of one of those three classes according in the AudioSet ontology.
This yields the following 110 audio classes:

Accordion;
Acoustic guitar;
Alto saxophone;
Bagpipes;
Banjo;
Bass (instrument role);
Bass drum;
Bass guitar;
Bassoon;
Bell;
Bicycle bell;
Bowed string instrument;
Brass instrument;
Bugle;
Cello;
Change ringing (campanology);
Chant;
Child singing;
Chime;
Choir;
Church bell;
Clarinet;
Clavinet;
Cornet;
Cowbell;
Crash cymbal;
Cymbal;
Dental drill, dentist's drill;
Didgeridoo;
Double bass;
Drill;
Drum;
Drum kit;
Drum machine;
Drum roll;
Electric guitar;
Electric piano;
Electronic organ;
Female singing;
Filing (rasp);
Flute;
French horn;
Glockenspiel;
Gong;
Guitar;
Hammer;
Hammond organ;
Harmonica;
Harp;
Harpsichord;
Hi-hat;
Jackhammer;
Jingle bell;
Keyboard (musical);
Male singing;
Mallet percussion;
Mandolin;
Mantra;
Maraca;
Marimba, xylophone;
Mellotron;
Musical ensemble;
Musical instrument;
Oboe;
Orchestra;
Organ;
Percussion;
Piano;
Pizzicato;
Plucked string instrument;
Power tool;
Rapping;
Rattle (instrument);
Rhodes piano;
Rimshot;
Sampler;
Sanding;
Sawing;
Saxophone;
Scratching (performance technique);
Shofar;
Singing;
Singing bowl;
Sitar;
Snare drum;
Soprano saxophone;
Steel guitar, slide guitar;
Steelpan;
String section;
Strum;
Synthesizer;
Synthetic singing;
Tabla;
Tambourine;
Tapping (guitar technique);
Theremin;
Timpani;
Tools;
Trombone;
Trumpet;
Tubular bells;
Tuning fork;
Ukulele;
Vibraphone;
Violin, fiddle;
Wind chime;
Wind instrument, woodwind instrument;
Wood block;
Yodeling;
Zither.

\subsection{Relevance}
\label{sec:relevance}

As described in Section~\ref{sec:retresults},
the AudioSet ontology is taken into account when evaluating the
retrieval performance, as, for example, an ideal system should rank
the `electric guitar` higher than `drums` when querying with an `acoustic guitar`.
We use the standard evaluation metric for this scenario where
retrieved results have varying relevance --
the normalized discounted cumulative gain (nDCG).
Here, we define the relevance between of one video to another.
Recall that AudioSet contains only video-level labels and that videos generally
have multiple labels.
Therefore, we first define the relevance of individual classes,
followed by the definition of the video (\ie set of classes) relevance.

\paragraph{Class relevance.}
An appropriate measure of distance between two classes organized in an ontology
is the tree distance, $d$, \ie the length of the shortest path between the
two classes. For example, the distances between `acoustic guitar` and
`acoustic guitar`, `electric guitar`, and `drums` are 0, 2 and 5, respectively.
The relevance of one class to another is then defined as the negative of their
tree distance, but offset by a constant to make sure relevances
are not negative.
Specifically, relevance is computed as $r=C-d$, where $C=20$ as this is
the longest possible distance between two classes.

\paragraph{Video relevance.}
Since videos generally contain multiple labels, we define the relevance of
one video to another as the maximal relevance across all pairs of
classes in the two videos.
The motivation behind using the maximal relevance,
as opposed to for example the minimal or the average,
is that AudioSet labels are only provided on the video-level.
Since we use only single frames or 1 second audio clips throughout,
it is not guaranteed that these contain all of the video classes
(in fact they could even contain none), so using a measure other than
the maximal relevance would over-penalize perfectly relevant results.
For example, consider the case of a video which has a person `singing`
followed by an `electric guitar`, and imagine we use a frame from
the second half of the video as a query.
The ground truth only tells us that there is `singing` and `electric guitar`
somewhere in the video, so we do not know which one of the two (if any)
does the frame depict. Therefore, retrieving a video which contains
`electric guitar` without `singing` is a perfectly acceptable result.

\section{Initialization for the AVE-Net}
In its vanilla form, there is actually nothing forcing the network
to make the distances between corresponding features small and
non-corresponding large -- it could equally learn anti-aligned embeddings
where a large distance between the visual and audio features signifies
high similarity. To stimulate the desired behaviour where small distance
means large similarity, one simply needs to enforce the correct sign of
the weights in the tiny \texttt{fc3} layer.
We found it to be sufficient to just initialize the layer with weights
of the correct sign and not enforce this during training.

\section{AVE+OF architecture}
\label{sec:aveof}

\figOFArch

Section 3.2 of the main paper discusses versions of the AVE-Net that
use multiple frames as input. Here we give details of the better performing
network, AVE+OF, which, along with a frame and 1 second of audio,
ingests 10 frames of optical flow as well
(computed using the TV-L1 algorithm~\cite{Zach07}).
The network follows the same architecture as the AVE-Net shown in
Figure 2d of the main paper, but with the vision subnetwork
(input: single RGB frame)
replaced with the network shown in Figure~\ref{fig:retarch:visionflow}
(input: single RGB frame and 10 optical flow frames).
The new vision subnetwork is a two-stream architecture~\cite{Simonyan14b},
\ie the frame and flow streams are fused by concatenation followed
by two convolutional layers.
The output of this network has the same dimensions as the original
vision ConvNet (Figure 2a of the main paper), and is therefore readily
pluggable into the AVE-Net architecture (Figure 2d of the main paper).

\section{Additional AVE-Net results}

Figures \ref{fig:ccadim} and \ref{fig:ndcgs} complement Section~\ref{sec:retresults},
and contain additional cross-modal retrieval results,
further demonstrating the superiority of AVE-Net versus all baselines.

\figDim
\figNdcg

}{}

\end{document}